\documentclass[letterpaper]{article} 
\usepackage{aaai2026}  
\usepackage{times}  
\usepackage{helvet}  
\usepackage{courier}  
\usepackage[hyphens]{url}  
\usepackage{graphicx} 
\usepackage{natbib}  
\usepackage{caption} 
\usepackage{algorithm}
\usepackage{algorithmic}

\usepackage{newfloat}
\usepackage{listings}
\usepackage{multirow}
\usepackage{amsmath,amssymb,amsfonts}
\usepackage{amsthm}
\newtheorem{prop}{Proposition}
\newtheorem{lemma}{Lemma}
\usepackage{booktabs}
\usepackage{soul}
\usepackage{xcolor} 
\sethlcolor{yellow}

\def\M{P_\theta}

\newtheorem{theorem}{Theorem}

\DeclareCaptionStyle{ruled}{labelfont=normalfont,labelsep=colon,strut=off} 
\floatstyle{ruled}
\newfloat{listing}{tb}{lst}{}
\floatname{listing}{Listing}
\title{Probabilistic Circuits for Knowledge Graph Completion with Reduced Rule Sets}
\author {
    Jaikrishna Manojkumar Patil\textsuperscript{\rm 1},
    Nathaniel Lee\textsuperscript{\rm 2},
    Al Mehdi Saadat Chowdhury\textsuperscript{\rm 2},
    YooJung Choi\textsuperscript{\rm 2},
    Paulo Shakarian\textsuperscript{\rm 1}
}
\affiliations {
    \textsuperscript{\rm 1}Syracuse University, Syracuse, NY USA\\
    \textsuperscript{\rm 2}Arizona State University, Tempe, AZ USA\\
    jpatil01@syr.edu, nlee51@asu.edu, achowd43@asu.edu, yj.choi@asu.edu, pashakar@syr.edu
}

\usepackage{bibentry}

\begin{document}

\maketitle

\begin{abstract}
    Rule-based methods for knowledge graph completion provide explainable results, but often require tens of thousands of rules to achieve competitive performance. Although individual predictions may use only a few rules, reasoning over an entire dataset requires these massive rule sets, hampering system-level understanding. We address this by learning a probability distribution over sets of rules that work together using probabilistic circuits. Our approach achieves a 70-96\% reduction in the number of rules needed to reach peak baseline performance. Using an equivalent minimal number of rules, we outperform the baseline by up to 31$\times$. When comparing our minimal rule sets against baseline's full rule sets, we preserve 91\% of peak baseline performance. Empirical validation on 8 benchmark datasets shows that our reduced rule sets exhibit higher utilization---fewer rules are wasted, and each prediction requires fewer rules. We show that our framework is grounded in well-known semantics of Nilsson's probabilistic logic and does not require independence assumptions. We provide exact probabilistic inference as well as an efficient lower bound and evaluate both. 
\end{abstract}

\section{Introduction}\label{sec:introduction}
Knowledge graph (KG) completion is a canonical problem in AI~\cite{hayes1977defence_logic,reiter1980logic_default_reasoning} which aims at inferring missing facts from incomplete data. It has applications spanning from search engines and recommendation systems to question answering systems and healthcare research. Recent developments~\cite{cai2026entity,yao2025exploring,yang2025gs,li2025multi,yue2025mlkgc,zhang2024making} have achieved remarkable improvements in this field. There are mainly two lines of work: high-performing but black box embedding-based approaches and explainable rule-based approaches. This research focuses on the latter. Rule-based methods such as AnyBURL~\cite{meilicke2024anytime} and AMIE+~\cite{galarraga2015fast_amieplus} learn first-order Horn rules from the KGs, providing explainable predictions through logical inference chains.
However, achieving this competitive performance with rule-based approaches leads to a fundamental challenge that undermines their core advantage. 
To reach peak performance on standard benchmarks, AnyBURL requires tens of thousands of rules. While each individual prediction may use only a small number of rules due to AnyBURL's max+ and noisy-or aggregation strategy, the overall rule set required to reason over an entire dataset remains significantly large. This creates three interconnected problems for system-level explainability.
First, Table~\ref{tab:pc_utilization_improvement} reveals that a substantial fraction of learned rules are never active during inference. For example, on the UMLS dataset, AnyBURL requires 20,000 rules to achieve Hits@10 = 0.9644, and only 12,938 rules (64.69\%) are actually used— leaving 7,062 wasted rules. 
This waste complicates system-level understanding because users cannot easily identify which rules actually contribute to the system's behavior versus which are irrelevant. Second, storing and managing tens of thousands of rules creates significant memory overhead. A large rule set with many inactive rules hides the reasoning patterns the system has learned. This memory issue is also acute if the results of the rule-learner are used as input for an LLM~\cite{bramblett2025discovering} which has limited context for rules. Third, reasoning tasks beyond simple prediction—such as consistency checking and abduction become computationally intractable as the search space grows with the number of rules. These tasks are the key components of trustworthy AI systems but become impractical with massive rule sets.

This raises a critical question: \textit{Can we achieve similar performance for KG completion tasks with rule-based approaches using a much smaller and coherent set of rules?}
\begin{table}[tb]
\small
\setlength{\tabcolsep}{3pt}
\centering
\scalebox{0.9}{
\begin{tabular}{lrrrrr}
\toprule
\textbf{Dataset} & \textbf{Hits@10} & \textbf{Total Rules} & \textbf{Active} & \textbf{Inactive} & \shortstack{\textbf{Avg \#rules}\\\textbf{per query}} \\
\midrule
\multicolumn{5}{c}{\textbf{Baseline (AnyBURL)}} \\
\midrule
UMLS & 0.964 & 20,000 & 12,938 & 7,062 & 58.50\\
CODEX-S & 0.426 & 20,000 & 12,218 & 7,782 & 15.89\\
WN18RR & 0.500 & 500 & 322 & 178 & 2.00\\
\midrule
\multicolumn{5}{c}{\textbf{PC Approach (this paper)}} \\
\midrule
UMLS & 0.964 & 1,000 & 868 & 132 & 25.80\\
CODEX-S & 0.426 & 1,000 & 642 & 358 & 1.70\\
WN18RR & 0.500 & 100 & 97 & 3 & 1.96\\
\bottomrule
\end{tabular}}
\caption{Rule Utilization Improvement. PC approach achieved comparable performance with significantly fewer total rules, higher utilization rates, and fewer inactive rules.}
\label{tab:pc_utilization_improvement}
\end{table}
To achieve this, we propose a framework that uses probabilistic circuits (PCs) to learn a distribution over sets of rules that work together from training data. By learning this distribution without independence assumptions between rules, we can identify smaller, high-utilization rule sets that achieve competitive performance while dramatically improving system-level explainability. 
For inference, we use either the lower bound of true probability or the exact probability to assign probabilities to predictions based on the rule sets that entailed them. We also provide a suite of formal results characterizing the computation of the query probability, and prove that our framework is equivalent to Nilsson's probabilistic logic~\cite{Nilsson1986ProbabilisticLogic}, grounding it in classical logic programming semantics.
Note that our framework is agnostic to the underlying rule sources and is not limited to AnyBURL, making it applicable to rules learned from any rule learning system.
We evaluated our framework on eight standard benchmark datasets.
We saw that the proposed framework achieved significant efficiency improvement. In these eight datasets, it reduces the size of the rule set by 70--96\% (an average 12-fold reduction) and delivers around 31-fold performance improvement over the confidence-based baseline. Additionally, this dramatic rule reduction still attains an average of 91\% of the highest baseline performance (i.e., baseline with maximum number of rules). Our reduced rule sets show higher utilization, i.e. fewer inactive rules during inference which directly improves system-level interpretability and thus reduces computational overhead for complex reasoning tasks.
While our primary focus in this paper is leveraging PCs for KG completion task, our framework has general applicability to any other domains of symbolic AI that struggle with scalability of rule-based approaches.\\
\noindent\textbf{Related Work.} \label{sec:related_work}

Embedding approaches such as CPE~\cite{liu2025contrastive}, NestE~\cite{xiong2024neste}, and RotatE~\cite{sun2019rotate}
have shown strong empirical performance for KG completion tasks, but are black-box methods and not the focus of this work. 
In contrast, rule-based approaches offer explainability and interpretability with frameworks like AnyBURL, NCRL~\cite{cheng2023neural},
RLogic~\cite{cheng2022rlogic},
RuleN~\cite{omran2018scalable_ruleN}, 
NeuralLP~\cite{yang2017differentiable_Nurallp}
capable of learning tens of thousands of logical rules from KGs. While these approaches address the scalability of rule learning systems, they still generate large rule sets that compromise explainability. Research works~\cite{zhang2020knowledge_kgc_pr,corte2018pilp} highlight the ongoing challenge of managing large number of learned rules. Current benchmark inference engines such as SAFRAN~\cite{ott2021safran}, and AnyBURL have achieved competitive performance with many embedding approaches across benchmark datasets. However, these approaches rely on confidence-based thresholding to select rule sets and require tens of thousands of rules to achieve competitive performance.
Recently, probabilistic circuits (PCs) have been used to integrate symbolic constraints with neural networks. 
Prior works like~\cite{ahmed2022nsyent,xu2018semanticloss} focus on the use of PC for constraint satisfaction on neural network outputs. Semantic Probabilistic Layers~\cite{NEURIPS2022_spl} uses PC to learn distributions in label configurations to ensure that the predictions satisfy the logical constraints. Recently~\cite{ahmed2024controllable} used probabilistic reasoning with constraint logic programs to guide text generation through Bayesian conditioning. The SLASH framework~\cite{skryagin2021slash} integrates PC into neural answer set programming for tractable symbolic reasoning guided by sub-symbolic structure. DeepProbLog~\cite{deepproblog2021} uses probabilistic logic programming that compiles programs into circuits for tractable inference, thus allowing neural networks to address predicate probabilities. 
Recently, \cite{AZZOLINI2025109497} learn mixtures over probabilistic logic programs. Under the distribution semantics, each component program defines a factorized distribution over rules, so a mixture of such programs corresponds to a shallow PC, while our framework instead learns a hierarchical PC over rule indicators and is rooted in Nilsson logic which does not assume independence between facts.
In contrast to all these approaches, our framework operates at rule selection level rather than prediction level, i.e. PC learns distribution over rule activations for scalable and highly transparent KG reasoning. 

\section{Technical Preliminaries}\label{sec:technical_prelims}
\noindent\textbf{Non-ground Syntax.} We assume a logic programming framework with a finite set of predicates and a finite set of constants. An \textit{atom} has the form $p(t_1, \ldots, t_k)$, where $p$ is a $k$-ary predicate and each $t_i$ is either a variable or a constant. A \textit{ground atom} is one in which every $t_i$ is a constant. A \textit{literal} is an atom or its negation. We define negation $\neg$ and conjunction $\wedge$ in the standard manner.

Following AnyBURL~\citep{meilicke2024anytime}, we work with binary rules of the form
\[
r:\quad h_r(X, Y) \;\leftarrow\; b_1(X, Z_1) \wedge b_2(Z_1, Z_2) \wedge \cdots \wedge b_n(Z_{n-1}, Y)
\]
where $X$ and $Y$ are the head variables and $Z_1, \ldots, Z_{n-1}$ are intermediate variables that chain the body atoms together. We write $b_r(X, Y)$ as shorthand for the entire body conjunction (with the intermediate variables left implicit). When the body is omitted ($h_r(X, Y) \leftarrow$), we call $r$ a \textit{fact}. A \textit{program} $\Pi$ is a set of rules. Note that our formal results do not depend on this specific form of rule. They hold for arbitrary rules of the form h $\leftarrow$ b where h is a literal and b is a conjunction of literals. We present the binary chain form above as it is the form produced by AnyBURL and used in our experiments.

For semantics, an \textit{interpretation} $w$ is a subset of ground atoms. Interpretation $w$ satisfies a ground atom if the atom is in $w$, satisfies the negation of a ground atom if the atom is not in $w$, and satisfies a conjunction of ground literals if each literal is satisfied. It satisfies a rule when, for every grounding of the rule's variables, satisfying the body implies satisfying the head. It satisfies a program $\Pi$ if it satisfies every rule in $\Pi$. Program $\Pi$ \textit{entails} a query $q$ (a program, rule, fact, conjunction, or literal) if for every interpretation $w$ with $w \models \Pi$ we have $w \models q$. We write this as $\Pi \models q$.

\noindent\textbf{Knowledge Graph and Triples.} A \textit{triple} is a ground atom of the form $p(c_1, c_2)$, where $p$ is a binary predicate and $c_1, c_2$ are constants. A \textit{knowledge graph} is a finite set of triples; we denote a fixed knowledge graph by $\Pi_G$, treating it as a program whose rules are all facts. The entailment relation from the previous subsection then specializes as follows: $\Pi_G \models p(c_1, c_2)$ iff the triple $p(c_1, c_2)$ is in $\Pi_G$, and $\Pi_G \models b_r(c_1, c_2)$ iff there exists a grounding of the intermediate variables $Z_1, \ldots, Z_{n-1}$ such that every body atom under that grounding is in $\Pi_G$. 

\noindent\textbf{Probability Distributions over Formulas.} For any ground formula $F$, $P_\theta(F)$ denotes the marginal probability that $F$ is true, with $P_\theta(\neg F) = 1 - P_\theta(F)$. The parameters $\theta$ are learned from data; we realize $P_\theta$ via a PC described next.

\noindent\textbf{Probabilistic Circuits.} We realize $P_\theta$ as a probabilistic circuit (PC)~\citep{choi2020probabilistic}, a computational graph representing a probability distribution over a set of variables. A PC is built from three types of nodes. The leaf nodes encode a distribution over a single variable. The product nodes multiply the distributions of their children over disjoint sets of variables and the sum nodes compute weighted mixtures of their children. Under standard structural properties (smoothness and decomposability), a PC supports exact computation of any marginal probability in time linear in the size of the circuit. This tractability is central to our framework as the inference methods we introduce later require repeated marginal queries over sets of Boolean variables each of which the PC answers efficiently. We also note that a fully factorized distribution is a PC with a single product node over its leaves and a flat mixture model is a PC with a single sum node. The general PCs are hierarchical mixtures and can therefore capture strictly richer families of distributions than either.

\section{Methodology}\label{sec:methodology}
\noindent\textbf{Rule Indicators and Samples.} Rules learned by data-mining algorithms like AnyBURL are not generally consistent with respect to a formal logical semantics, so we model rule activations probabilistically. For each rule $r \in \Pi$, we introduce a zero-arity indicator atom $\mu_r()$ representing whether $r$ is active. We rewrite each rule as
\[
h_r(X, Y) \leftarrow b_r(X, Y) \wedge \mu_r(),
\]
so that $\mu_r()$ acts as a switch, $r$ fires only when $\mu_r()$ is true.

A \textit{sample} is a training triple, i.e. a ground atom $p(c_1, c_2) \in \Pi_G$. For sample $s$ given by the triple $p(c_1, c_2)$, we introduce a zero-arity helper atom $\nu_s()$. We record what a sample tells us about each rule's activation as follows:
\begin{itemize}
    \item $\Pi_G \models b_r(c_1, c_2)$ and $h_r(c_1, c_2) = p(c_1, c_2)$, the KG satisfies the rule's body and its head matches the sample. We include $\mu_r() \leftarrow \nu_s()$.
    \item Otherwise, nothing is included; $\mu_r()$ is \textit{unobserved} for $s$.
\end{itemize}
As we describe below, this missing information is precisely what motivates our choice of learning.

\noindent\textbf{Association Matrix. }The rule-sample associations are determined via PyClause~\citep{betz2024pyclause}, which computes for each training triple, the set of rules that predict it~\cite{betz2022adversarial}. The associations are stored as a matrix $M$ where $M_{r,s} = 1$ if $\mu_r() \leftarrow \nu_s()$ is included and is missing otherwise. The set of all samples is denoted $S$, and we write $\Pi \cup \{\nu_s() \leftarrow\}$ as $\Pi_s$.

For the formal semantic results (Theorem~\ref{theorem:nilson}), distinct samples must have mutually exclusive helper atoms; this is captured by the rules $\neg \nu_{s'}() \leftarrow \nu_s()$ for every pair $s' \neq s$. In implementation, we do not actually add these quadratically many rules; instead, we reason with each $\Pi_s$ in isolation, which gives the same mutual-exclusivity guarantee.


\noindent\textbf{Learned Probability Distribution over Rule Indicators.} Given the association matrix $M$, we treat each sample as one observation of an underlying joint distribution over rule indicators and use maximum-likelihood learning to fit the PC $P_\theta$ over these indicators. PC learning has two phases: \textit{structure learning} identifies a circuit structure suited for the dependencies between rule indicators, and \textit{parameter learning} fits the leaf distributions and sum-node weights via expectation maximization (EM). EM is important here because the matrix has missing entries, i.e. for any pair $(r, s)$ where $\mu_r()$ is unobserved, EM treats $\mu_r()$ as a latent variable and fits parameters that maximize the likelihood of the observed data. 

To see why a general joint distribution matters, consider a fully factorized distribution over rule indicators:
\begin{equation}
    P_{fact}(R) = \prod_{r \in R} P_\theta(\mu_r()).
\end{equation}
Here, each rule indicator is treated as independent of every other. This is the implicit model behind traditional confidence-based rule scoring, where each rule is assigned an isolated probability. However, rule indicators are typically not independent and usually show correlations in their activations. The PC instead learns a general (non-factorized) joint distribution over rule indicators. The marginal $P_\theta(\mu_r())$ under the PC depends not only on how often $r$'s indicator is observed active, but also on how its activation correlates with the activations of other rules.

The formal results in the next section are stated in terms of a probability mass $P_\theta(s)$ associated with each sample. This is defined through the helper atoms: $P_\theta(s)$ denotes the probability that $\nu_s()$ holds and by mutual exclusivity of the helper atoms, distinct samples contribute disjoint mass, so $\sum_s P_\theta(s) = 1$. The sample-level and indicator-level views are connected by:
\begin{equation}
    P_\theta(R) = \sum_{s\;:\;\Pi_s \models R} P_\theta(s).
\end{equation}
That is, the marginal probability of a rule set equals the total mass of the samples whose programs entail it. Note that $\Pi_s \models R$ cannot be determined for every pair of $s$ and $R$, i.e. when $\mu_r()$ is unobserved for $s$, sample $s$ carries no information about whether $r$ is active. This hidden information is what EM resolves during learning, and Theorem~\ref{theorem:nilson} later establishes that the resulting distribution is consistent with formal probabilistic semantics.
Throughout the paper, we use $q$ to denote a query (a program, rule, fact, conjunction, or literal) and $R$ to denote a set of rules. The learning process is summarized in Algorithm~\ref{alg:learn_pc} (line-by-line walkthrough in the Appendix).

\begin{algorithm}[tb]
    \caption{Learning Probabilistic Circuit}
    \label{alg:learn_pc}
    \textbf{Input}: $\Pi$, $T_{\text{train}}$ (training triples)\\
    \textbf{Output}: $P_\theta$ (learned PC over rule indicators)
    \begin{algorithmic}[1]
        \STATE $M \gets \textsc{ConstructAssociationMatrix}(\Pi, T_{\text{train}})$ 
        \STATE $\pi \gets \textsc{LearnPCStructure}(M)$
        \STATE $\theta \gets \textsc{LearnPCParameters}(M, \pi)$
        \RETURN $P_\theta$
    \end{algorithmic}
\end{algorithm}

\section{Formal Results}\label{sec:formal_results}

In this paper, we shall often find it computationally advantageous to bound or find equivalences to the marginal. We also take a simplified view that each training triple is a sample. This choice enables efficient computation while maintaining the formal semantics described here. In this section, we prove a suite of results for this reason. None of the following results rely on any sort of independence assumption.  We first show a lower bound on the marginal. 
\begin{prop}
\label{prop:lb1}
For any set of rules $R$ and a sample $s$,
\begin{eqnarray}
\M(R) \geq \sum_{s \textit{ s.t. }R \subseteq \Pi_s}P_\theta(s)
\end{eqnarray}
\end{prop}

This proposition follows from the fact that if $R$ is contained in $\Pi_s$ then $R$ must be a logical consequence.

We next look at the marginal of specific query. We assume that the query is \textit{intensional}, meaning that it does not appear in any facts of $\Pi$ but can appear in the head of non-fact rules. 

\begin{prop}
\label{prop:eq}
When $q$ is intensional,
\begin{eqnarray}
\M(q) = 1-\M\left(\bigwedge_{s \textit{ s.t. } \Pi_s \models q}\neg \nu_s()\right)
\end{eqnarray}
\end{prop}

We can also show properties to relate marginal probabilities to each other. This can be useful if we desire to only compute the marginal probability of a small number of rule sets. This is particularly useful when the number of samples (the size of $S$) is very large (e.g., in our experiments we learned probability distributions with $10^5$ samples). Further, while it is advantageous to learn $P_\theta$ over large numbers of program for reasons of accuracy, this is detrimental to explainability when considering a user.

\begin{prop}
\label{prop:lb_prop}
Given subsets of $\Pi$: $R_1,\ldots,R_j,\ldots,R_{max}$,
\begin{eqnarray}
\M(q) \geq sup\{\M(R_j) | R_j \models q\}
\end{eqnarray}
\end{prop}

The corresponding upper bound mirrors the above proposition.

\begin{prop}
\label{prop:ub_prop}
Given subsets of $\Pi$: $R_1,\ldots,R_j,\ldots,R_{max}$,
\begin{eqnarray}
\M(q) \leq sup\{1-\M(R_j) | R_j \not\models q\}
\end{eqnarray}
\end{prop}

\noindent\textbf{Inference Procedure.} As we learn $P_\theta$ over the set of samples, if we have facts that are known to be true, then we can simply add those facts to $\Pi$ and, for a given query, we can compute its probability based on the facts and learned probability distribution by using computation techniques shown in Propositions \ref{prop:lb1}-\ref{prop:ub_prop}. We note that propositions \ref{prop:lb_prop} and \ref{prop:ub_prop} require the identification of some subsets of rules to enable the calculation of the marginals. We performed experiments with three such methods:
\begin{itemize}
\item \textbf{\underline{SingletonLB}.} Approximates the probability of a query with the lower bound of Proposition~\ref{prop:lb_prop} and uses rule sets consisting of singleton rules.
\item \textbf{\underline{SingletonExact}.} Exactly computes the probability of a query using Proposition~\ref{prop:eq}.
\item \textbf{\underline{GreedyLB}.} Approximates the probability of a query with the lower bound of Proposition~\ref{prop:lb_prop} and uses rule sets created from a greedy walk (more details in the Appendix). The walk is terminated once the marginal of the rule is below the predefined threshold. The new walk begins with the remaining non-ground rules.
\end{itemize}
Algorithm~\ref{alg:inference} describes the complete inference procedure, which works across all three variants (line-by-line walkthrough in the Appendix). 

\begin{algorithm}[tb]
    \caption{PC-Guided Inference}
    \label{alg:inference}
    \textbf{Input}: $\Pi$, $P_\theta$, $T_{\text{train}}$, $variant \in \{\text{SingletonLB}, \text{SingletonExact}, \text{GreedyLB}\}$, $r_{max}$\\
    \textbf{Output}: $Q$ (set of predicted triples)
    \begin{algorithmic}[1]
        \STATE Sort $\Pi$ by marginal probability: $r_1, r_2, \ldots, r_{|\Pi|}$ such that $P_\theta(\mu_{r_1}()) \geq P_\theta(\mu_{r_2}()) \geq \cdots \geq P_\theta(\mu_{r_{|\Pi|}}())$
        \STATE $\mathcal{S} = \emptyset$
        \IF{$variant \in \{\text{SingletonLB}, \text{SingletonExact}\}$}
            \FOR{$i = 1$ to $r_{max}$}
                \STATE $R \gets \{r_i\}$
                \STATE $\mathcal{S} \gets \mathcal{S} \cup (R, P_\theta(R))$
            \ENDFOR
        \ELSIF{$variant \in \{\text{GreedyLB}\}$}
            \STATE $\mathcal{S} \gets \textsc{GreedyWalk}(\Pi, P_\theta, r_{max})$
        \ENDIF
        \STATE $Q \gets \emptyset$
        \FOR{each $(R, \cdot) \in \mathcal{S}$}
            \STATE $Q \gets Q \cup \textsc{AnyBURLInference}(R, T_{\text{train}})$
        \ENDFOR
        \FOR{each $q \in Q$}
            \STATE $F_q \gets \{r \in \Pi : r \text{ active for } q\}$
            \STATE $P_\theta(q) \gets \textsc{ComputeTripleProb}(F_q, variant, P_\theta)$
        \ENDFOR
        \RETURN $Q$
    \end{algorithmic}
\end{algorithm}
\noindent\textbf{Formal Probabilistic Semantics.} Here we show that the presented formalism can be shown to be equivalent to Nilsson Logic~\cite{Nilsson1986ProbabilisticLogic} providing a formal probabilistic semantics. Let us consider the following construction in Nilsson logic, creating a logic program $\Pi_{nil}$ and $P_{nil}$ (where $P_{nil}$ assigns a probability to all elements of $\Pi_{nil}$):

\begin{enumerate}
\item For each $r \in \Pi$, add $r$ to $\Pi_{nil}$ and set $P_{nil}(r)=1.0$.
\item For each $s \in S$ add fact $\nu_s() \leftarrow $ to $\Pi_{nil}$ and set\\ $P_{nil}(\nu_s() \leftarrow) = \M(s)$.
\end{enumerate}

Following~\cite{Khuller2007Computing,Shakarian2012APT}, an interpretation $I$ is a probability distribution over all possible worlds - which is the same semantic structure of~\cite{Nilsson1986ProbabilisticLogic}. Given a logic program and associated probabilities assigned to all its elements (e.g., as in $\Pi_{nil}$ and $P_{nil}$ defined above), an interpretation $I$ satisfies the structure (written $I \models (\Pi_{nil},P_{nil})$ if for all $r \in \Pi_{nil}$, we have the following:
\begin{eqnarray}
\sum_{w \textit{ s.t. }w\models r}I(w) = P_{nil}(r)
\end{eqnarray}
With this in mind, we first prove a technical lemma.
\begin{lemma}\label{lemma:helper}
For some formula $q$, and world $w$ where a satisfying interpretation $I$ assigns it a non-zero probability, $w \models q$ if and only if for some $s$ where $\Pi_s \models q$, $w \models \nu_s()$.
\end{lemma}
\noindent\textit{(Proof in Appendix.)}

Therefore, we have the following result to show that our syntactically-learned probability distribution ensures semantic entailment:
\begin{theorem}\label{theorem:nilson}
Interpretation $I$ that satisfies $(\Pi_{nil},P_{nil})$ if and only if for any query formula $q$:
\[
\sum_{w \textit{ s.t. }w\models q}I(w) = \sum_{s \textit{ s.t. }\Pi_s \models q}P_\theta(s)
\]
\end{theorem}
\noindent\textit{(Proof in Appendix.)}

\section{Experiments}\label{sec:experiments}

We empirically evaluate our PC-guided inference framework on multiple KGs, comparing its performance with a standard confidence-based inference baseline.
\begin{table}[htb]
\small
\setlength{\tabcolsep}{2pt} 
\centering
\begin{tabular}{lrrrrr}
\toprule
\textbf{Dataset} & \textbf{\#Entities} & \textbf{\#Relations} & \textbf{\#Train} & \textbf{\#Test} \\
\midrule
FB15K-237 & 14,541 & 237 & 272,115 & 20,466 \\
WN18      & 40,943 & 18  & 141,442 & 5,000  \\
WN18RR    & 40,943 & 11  & 86,835  & 3,134  \\
CODEX-S   & 2,034  & 42  & 32,888  & 1,828  \\
Kinship   & 104    & 26  & 8,544   & 1,074  \\
Family    & 2,411  & 12  & 5,868   & 2,835  \\
UMLS      & 135    & 49  & 5,216   & 661    \\
Nations   & 14     & 56  & 1,592   & 201    \\
\bottomrule
\end{tabular}
\caption{Statistics of benchmark datasets for KG Completion.}
\label{tab:dataset-stats}
\end{table}
\begin{figure*}[ht!]
\centering
\begin{tabular}{@{}c@{\hspace{0em}}c@{\hspace{0em}}c@{\hspace{0em}}c@{}}

\begin{tabular}{@{}c@{}}
    \textbf{WN18RR} \\[0em]
        \includegraphics[width=0.25\linewidth]{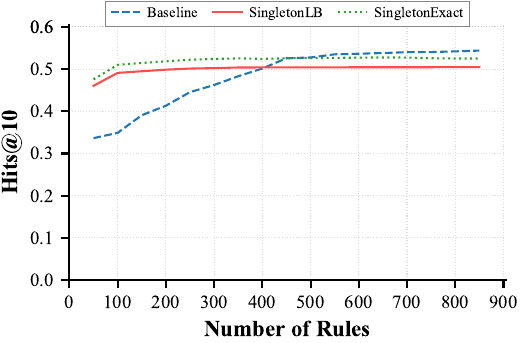}\\[0.8em]

    \textbf{WN18} \\[0em]
        \includegraphics[width=0.25\linewidth]{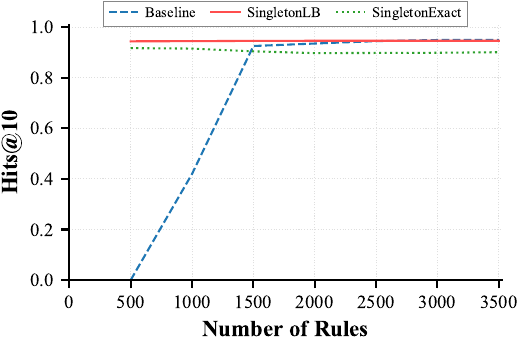}
\end{tabular}
&

\begin{tabular}{@{}c@{}}
    \textbf{Nations} \\[0em]
        \includegraphics[width=0.25\linewidth]{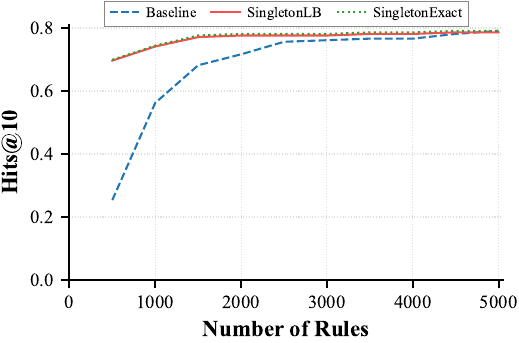}\\[0.8em]

    \textbf{Family} \\[0em]
        \includegraphics[width=0.25\linewidth]{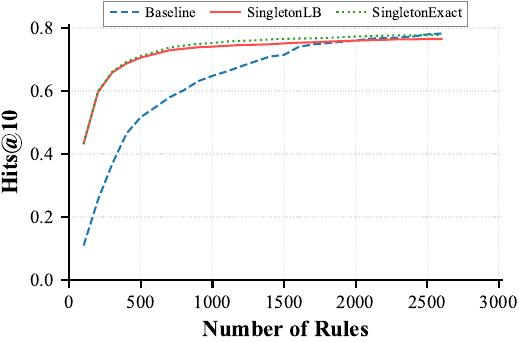}
\end{tabular}
&

\begin{tabular}{@{}c@{}}
    \textbf{FB15K-237} \\[0em]
        \includegraphics[width=0.25\linewidth]{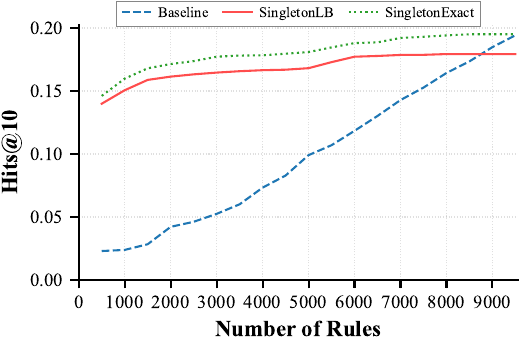}\\[0.8em]

    \textbf{Kinship} \\[0em]
        \includegraphics[width=0.25\linewidth]{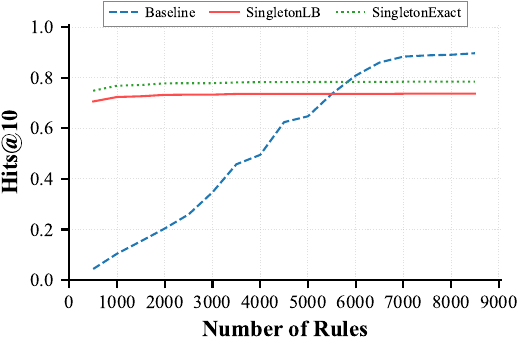}
\end{tabular}

&

\begin{tabular}{@{}c@{}}
    \textbf{CODEX-S} \\[0em]
        \includegraphics[width=0.25\linewidth]{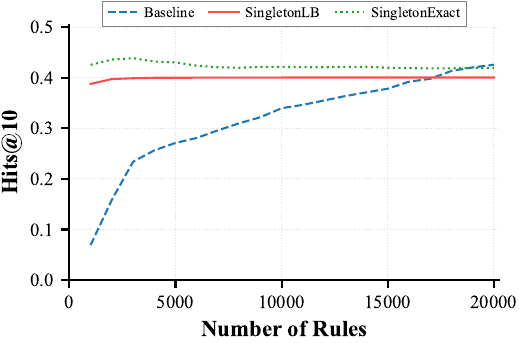}\\[0.8em]

    \textbf{UMLS} \\[0em]
        \includegraphics[width=0.25\linewidth]{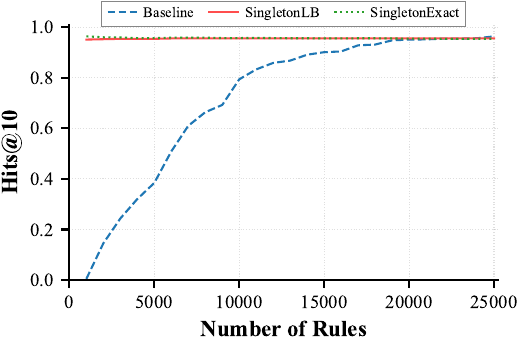}
\end{tabular}

\end{tabular}

\caption{Hits@10 performance of SingletonLB and SingletonExact inferences against Baseline. PC variants attained competitive performance for Hits@10 with reduced rule sets across all evaluated datasets.}
\label{fig:hits_at_10}
\end{figure*}
\begin{figure*}[ht!]
\centering
\begin{tabular}{@{}c@{\hspace{0em}}c@{\hspace{0em}}c@{\hspace{0em}}c@{}}

\begin{tabular}{@{}c@{}}
    \textbf{WN18RR} \\[0em]
        \includegraphics[width=0.25\linewidth]{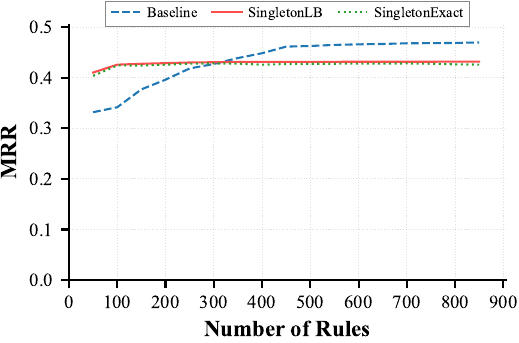}\\[0.8em]

    \textbf{WN18} \\[0em]
        \includegraphics[width=0.25\linewidth]{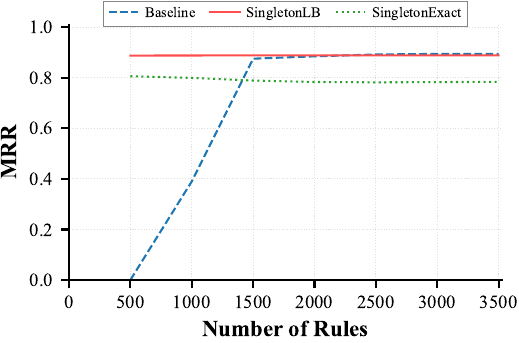}
\end{tabular}
&

\begin{tabular}{@{}c@{}}
    \textbf{Nations} \\[0em]
        \includegraphics[width=0.25\linewidth]{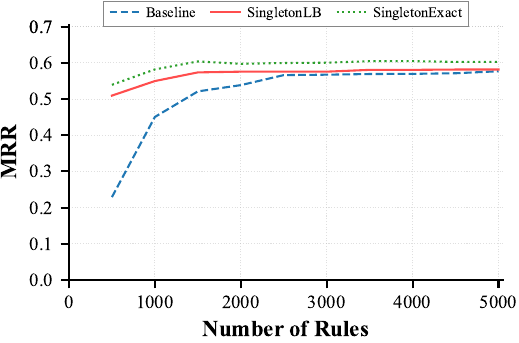}\\[0.8em]

    \textbf{Family} \\[0em]
        \includegraphics[width=0.25\linewidth]{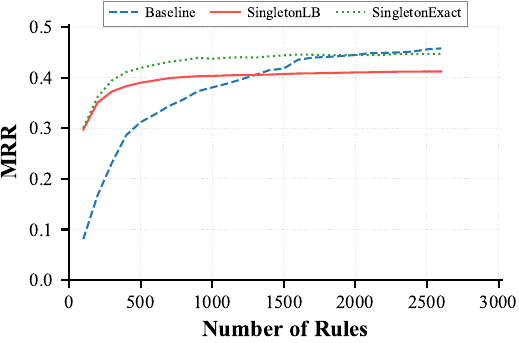}
\end{tabular}
&

\begin{tabular}{@{}c@{}}
    \textbf{FB15K-237} \\[0em]
        \includegraphics[width=0.25\linewidth]{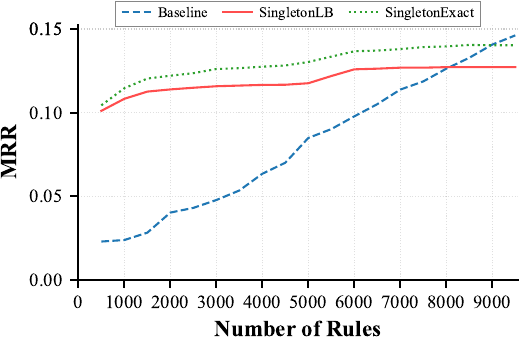}\\[0.8em]

    \textbf{Kinship} \\[0em]
        \includegraphics[width=0.25\linewidth]{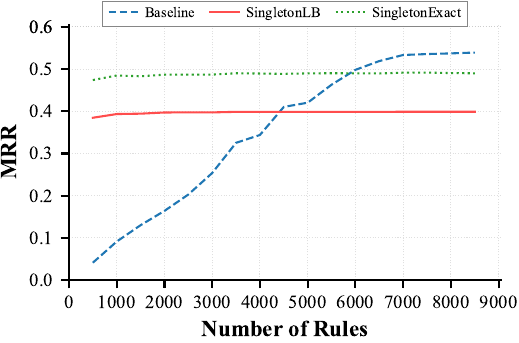}
\end{tabular}

&

\begin{tabular}{@{}c@{}}
    \textbf{CODEX-S} \\[0em]
        \includegraphics[width=0.25\linewidth]{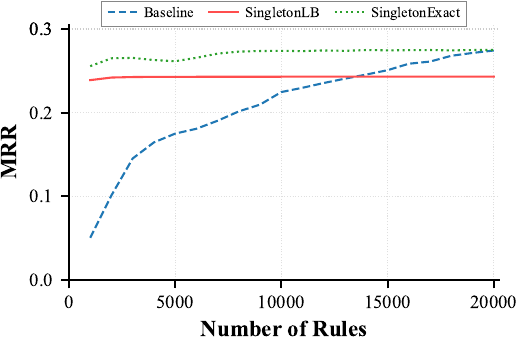}\\[0.8em]

    \textbf{UMLS} \\[0em]
        \includegraphics[width=0.25\linewidth]{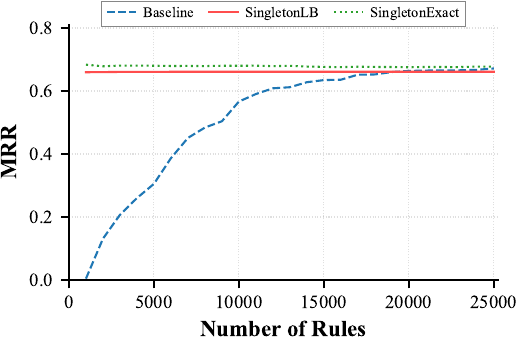}
\end{tabular}
\end{tabular}
\caption{Mean rank reciprocal (MRR) performance of SingletonLB and SingletonExact inferences against Baseline. PC variants attained competitive performance for MRR with reduced rule sets across all evaluated datasets.}
\label{fig:mrr}
\end{figure*}\\
\noindent\textbf{Datasets.}
We evaluate our PC-guided inference framework on 8 standard benchmark datasets: FB15K-237~\cite{toutanova2015representing}, WN18RR~\cite{wn18rr}, WN18~\cite{bordes2013translating}, Family~\cite{cheng2023neural}, Nations~\cite{kok2007}, UMLS~\cite{kok2007}, CODEX-S~\cite{safavi-koutra-2020-codex}, and Kinship~\cite{kok2007}. 
Additionally, these datasets together allow comprehensive evaluation over different characteristics like dataset size, domain (general knowledge, biomedical, family relations), and rule learning complexity based on number of entities and relations. Table~\ref{tab:dataset-stats} provides the key statistics for all the datasets used in the experiments.\\
\noindent\textbf{Experimental Setup.}
Our evaluation task is KG completion: given a KG and a set of learned non-ground rules, the goal is to infer missing triples that are likely to be in test set. We use standard metrics such as Hits@1, Hits@3, Hits@10 and Mean Rank Reciprocal (MRR) to compute the quality of predicted triples. 
Hits@k gives the proportion of test triples that appear in top-k predictions, and MRR computes the average reciprocal rank of correct predictions across all test triples (formal definition of both in Appendix). All reported metrics are filtered: the AnyBURL evaluation engine computes filtered Hits@k and a top-k lower bound of filtered MRR, consistent with standard KGC evaluation practice.
We use AnyBURL rule learner with 10 seconds learning time and minimum support threshold $\ge 10$ for consistent comparisons. AnyBURL assigns a confidence score to each of these learned rules as the ratio of instances where both the rule's head and body were satisfied to the number of instances where only the body was satisfied.
Each training triple serves as a sample, and the rule-sample association matrix $M$ is constructed via PyClause as described in the Methodology section. Note that we do not represent the infrastructure rules  $\mu_r() \leftarrow \nu_s()$ explicitly but instead rely on the association matrix.
For PC learning, we use the Hidden Chow-Liu Tree~\cite{liu2021tractable_hclt} for structure learning and expectation maximization (EM) for parameter learning using the Juice library for PCs~\cite{choi2020probabilistic}. 
Note that 5 runs of the framework were conducted for Family dataset to check stability of results, rest were run once.
\\
\noindent\textit{Hyperparameter Selection.}
We tested confidence thresholds for input non-ground rules (50--90\%) and number of EM iterations for circuit parameter learning (10, 50, 100). The final settings focused on showing framework effectiveness over large number of rules for each dataset and also ensuring computational tractability for evaluation. The confidence thresholds for each evaluated dataset are 50\% (CODEX-S/Kinship/UMLS), 60\% (FB15K-237), 70\% (Nations), 0\% (all mined rules for WN18, WN18RR, Family). Datasets with higher thresholds (e.g., Kinship at 90\% with only 1300 rules) outperformed respective baseline too but does not prove scalability to larger rule base. The number of EM iterations are 100 (Family/WN18/WN18RR), 50(Kinship), 10(Others).\\
\noindent\textit{Baseline/PC Inference.}
Baseline sorts all the input rules by confidence and selects top-$n$ rules for evaluation point $n$ and infers using AnyBURL inference engine. Our PC-guided framework uses the SingletonLB, SingletonExact, and GreedyLB inference methods described in the Formal Results section to obtain the prediction probabilities. AnyBURL evaluation engine was used to compute Hits@k and MRR for all methods.
\section{Results and Discussion}

All experiments were conducted on high-throughput computing cluster nodes utilizing up to 16-core CPUs (single cores for simpler tasks), NVIDIA A100 GPU, and up to 500 GB memory. Note that 500 GB is the memory available on the cluster nodes, not the amount consumed. The actual memory consumption scales linearly with the number of rules.
We empirically evaluated our framework with eight benchmark knowledge graphs and demonstrated that our framework maintained competitive performance with much higher efficiency (in terms of the required number of rules to reach there) than the baseline approach. 
For these 8 datasets, our framework reduced the required size of rule set between 70-96\% (average 12-fold reduction in rules). With this reduced rule count, we outperform the baseline by approximately $31\times$. Our approach also preserved an average of around 91\% of highest baseline performance (PC-guided performance with minimal rules vs. baseline's confidence-based performance with maximum considered rules).

\noindent\textbf{Hits@10. }In Figure~\ref{fig:hits_at_10}, we see consistent patterns of immediate performance for Hits@10 over all the datasets. Using minimal equivalent number of rules, SingletonExact inference shows substantial improvements for Hits@10 over the baseline (with SingletonLB showing similar trends): 214-fold (UMLS), 17-fold (Kinship), 6.4-fold (FB15K-237), 6.2-fold (CODEX-S), 4-fold (Family), 2.8-fold (Nations), 1.4-fold (WN18RR) and significant improvement in WN18 as baseline achieved zero performance for 500 rules. 
In CODEX-S, SingletonExact inference achieved Hits@10 of 0.4259 using only 1000 rules by achieving 99.95\% of the baseline's highest Hits@10 (0.4261) that required 20000 rules, i.e. SingletonExact required only 5\% of the total rules to reach there. UMLS gives the strongest evidence for the need of our framework where SingletonExact inference achieves baseline's peak Hits@10 of 0.9644 by using only 1000 rules compared to baseline that required 25000 rules to reach. This is a 25-fold reduction in rule usage. In WN18, we see both SingletonLB and SingletonExact inferences achieving near-peak performance(99.4\% and 99.22\%) of baseline's using only 500 rules. The baseline struggled from zero performance before reaching its peak at 1500 rules. This highlights our framework's ability to ignore these insignificant high-confidence rules. Nations and Family show superiority in performance by PC approaches with minimal rule requirements by achieving 295\% and 175\% performance gains respectively when comparing with identical rule counts (500 for Nations and 100 for Family). Finally, FB15K-237 illustrates scalability to larger datasets. We see that SingletonExact reached 87\% of the highest baseline performance with 85\% fewer rules. We also see improvement in rule utilization in Table~\ref{tab:pc_utilization_improvement} for highest Hits@10 for SingletonExact. For example, our framework had 89.6\% active rules out of only 500 total rules to reach the highest Hits@10 compared to baseline which had only 38\% active rules out of total 1500 rules for WN18.

\noindent\textbf{MRR.} Figure~\ref{fig:mrr} reveals consistent superior MRR performance over baseline across all datasets when using equivalent minimal rules: 151-fold (UMLS), 11-fold (Kinship), 4.5-fold (FB15K-237), 5-fold (CODEX-S), 4-fold (Family), 2.4-fold (Nations), 1.2-fold (WN18RR) and significant improvement in WN18 where baseline achieved zero performance for 500 rules. 
Beyond this comparison for minimal number of rules, the analysis also confirms that our framework not only identifies correct predictions but also ranks the predictions with high quality probabilistic inference. In UMLS and Nations, SingletonExact inference outperformed the highest baseline (baseline uses maximum available rules) by 1.78\% and 4.7\% respectively when SingletonExact used 96\% and 70\% less number of rules for each dataset. WN18 uniquely showed SingletonLB MRR outperforming SingletonExact MRR using 500 rules, but both SingletonLB and SingletonExact outperform the highest baseline with 7$\times$ reduction in rules. For FB15K-237, WN18RR, and Kinship, SingletonExact achieved 82.35\%, 90.34\%, and 88\% of best baseline MRR with around 89\% less number of rules (similar Hits@1 and Hits@3 results in Appendix).
\begin{figure}[htbp]
\centering
\includegraphics[width=0.49\columnwidth]{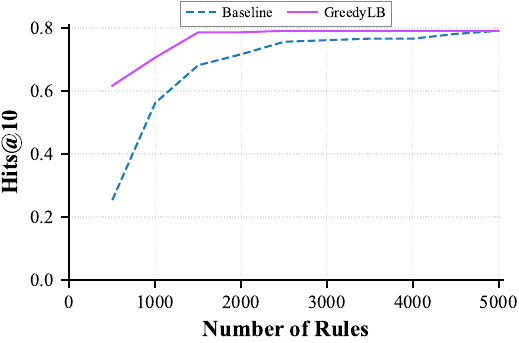} 
\includegraphics[width=0.49\columnwidth]{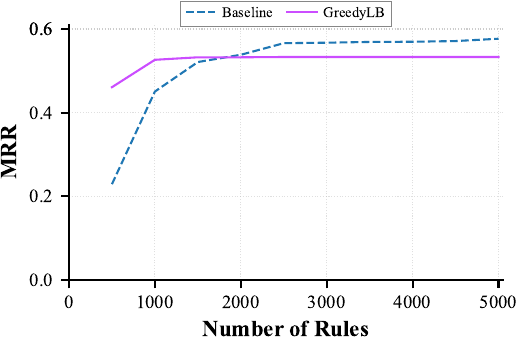} 
\caption{Greedy walks performance on Nations dataset. Comparison of GreedyLB and Baseline for Hits@10 (left) and MRR (right).}
\label{fig:nations_hits_mrr_greedy}
\end{figure}
\\
\noindent\textbf{Results using GreedyLB inference method (Greedy walks).}
Figure~\ref{fig:nations_hits_mrr_greedy} compares GreedyLB with the baseline for the Nations dataset. GreedyLB outperforms the baseline with at least $2\times$ better Hits@10 and MRR using 500 rules, showing that all three PC-guided inferences outperform the baseline for comparison with minimal rulesets (GreedyLB results for more datasets in the Appendix). However, comparing with Figure~\ref{fig:hits_at_10} and Figure~\ref{fig:mrr} reveals the superiority of SingletonLB and SingletonExact over GreedyLB. SingletonLB achieves 13\% higher Hits@10 and 11\% higher MRR for 500 rules. GreedyLB's Hits@10 eventually catches up to SingletonLB at 1500 rules, but MRR remains consistently lower. Additionally, computational overhead for greedy walks adds unnecessary complexity (details in Appendix). 
This shows that computational sophistication did not lead to superior performance.

\noindent\textbf{Runtime analysis.}
Our framework scales linearly with the number of input rules. The runtime was consistently linear across training as well as inference: generating rule samples ($R^2=0.90$), learning the PC ($R^2=0.99$), singleton rule sets ($R^2=0.91$), SingletonLB inference ($R^2=0.96$), and SingletonExact inference ($R^2=0.95$). Circuit learning during training takes most of the computation time but enables improved explainability. Full results on runtime analysis are included in the Appendix.
\section{Conclusion and Future Work}\label{sec:conclusion}
The embedding-based approaches have shown state-of-the-art performances for KG completion, but this comes at the cost of interpretability. In recent years, explainable rule-based methods have achieved competitive performances over embedding-based approaches. However, this requires grounding tens of thousands of non-ground rules, undermining the very explainability they were designed for. In this paper, we address the critical challenge of preserving reasoning performance while significantly reducing rule sets for interpretable inference. 
We proposed a PC guided framework that identifies rule–sample associations from training data and learns a joint distribution over rule activations without assuming rule independence. We also provided a suite of formal results characterizing the computation of probability and showing the adherence to formal probabilistic semantics.
We introduced 3 PC-guided inference methods: SingletonLB, SingletonExact, and GreedyLB. Our framework reduced the rule requirement by 70-96\%, and with these smaller rule sets, we got 31-fold performance (Hits@10, MRR) improvement over baseline with same number of baseline rules. We also preserve an average of 91\% performance with a reduced rule set compared to the full rule set of the baseline.

\noindent\textbf{Future Work. }
As our framework does not depend on any graph-specific data structures, it can be applied to other symbolic AI domains that struggle with rule explosion such as proof search in symbolic theorem proving~\cite{jiang2024leanreasoner}. 
As our framework is agnostic to the underlying rule sources, 
we will explore the integration of the framework with other rule learning systems~\cite{cheng2023neural,safelearner,de2015inducingprobfoil_plus}, particularly Inductive logic Programming systems to address their exponential search space challenges~\cite{corte2018pilp}. We would also like to investigate how our framework can improve neuro-symbolic systems~\cite{weber2019nlprolog} by providing reduced rule sets alongside sub-symbolic components.

\section{Acknowledgments}
Funded by ARO grant W911NF-24-1-0007.
\bibliography{aaai2026}

\appendix
\newpage
\section{Proofs}
\noindent\textit{Proof of Proposition~\ref{prop:eq}.}
\begin{proof}
\begin{eqnarray}
\M(\bigwedge_{s \textit{ s.t. } \Pi_s \models q}\neg \nu_s()) = \sum_{s' \textit{ s.t. }\Pi_{s'} \models \bigwedge_{s \textit{ s.t. } \Pi_s \models q}\neg \nu_s()}P_\theta(s')
\end{eqnarray}
Consider some $\Pi_{s''}$. If $\Pi_{s''} \models q$ then it does not model $\bigwedge_{s \textit{ s.t. } \Pi_s \models q}\neg \nu_s()$ (as the fact $\nu_{s''}()\leftarrow$ is in $\Pi_{s''}$). If it does not model $q$, then for every other $s'''$ the rule $\neg \nu_{s'''}() \leftarrow \nu_{s''}()$ implies $\Pi_{s''}\models \neg \nu_{s'''}()$. Therefore, in this case $\Pi_{s''}\models \bigwedge_{s \textit{ s.t. } \Pi_s \models q}\neg \nu_s()$. Therefore, we obtain the following equivalence:
\begin{eqnarray}
\{s' \textit{ s.t. }\Pi_{s'} \models \bigwedge_{s \textit{ s.t. } \Pi_s \models q}\neg \nu_s()\}=\{s' \textit{ s.t. }\Pi_{s'} \not\models q\}
\end{eqnarray}
From this, we get the following:
\begin{eqnarray}
\M(\bigwedge_{s \textit{ s.t. } \Pi_s \models q}\neg \nu_s()) &=&\sum_{s \textit{ s.t. }\Pi_{s} \not\models q}\M(s)\\
&=&1-\sum_{s \textit{ s.t. }\Pi_{s} \models q}\M(s)\\
&=&1-\M(q)
\end{eqnarray}
Which in turn gives us the statement.
\end{proof}

\noindent\textit{Proof of Lemma~\ref{lemma:helper}.}
\begin{proof}

\noindent Claim 1: Any satisfying interpretation that assigns a world $w$ a non-zero probability must satisfy exactly one $\nu_s()$ atom. We note that this follows from the fact that, by the construction, no two $\nu$ atoms can be true in a given world (by the construction) and as they are associated with disjoint samples, the probability mass assigned any single $\nu$ atom must sum to $1$ (by $P_{nil}$).\\

\noindent Claim 2: ($\rightarrow$) Suppose, BWOC, that $w \models q$ and there is no $s$ where $\Pi_s \models q$, $w \models \nu_s()$. Claim 1 tells us that $w$ must satisfy a single helper atom - let us call this $\nu_s()$. Consider the rules of the form $\mu_r() \leftarrow \nu_s()$ and $h_r \leftarrow b_r \wedge \mu_r()$ in $\Pi_{nil}$. As all other rules cannot possibly fire as only $\nu_s()$ is true in $w$ (which by Claim 1 precludes other helper atoms), and the set of rules of second form can fire, and only these rules entail $q$. Note these are the same rules in $\Pi_s$. Hence, these rules together must entail $q$. This tells us that $\Pi_s \models q$ - a contradiction.\\

\noindent Claim 3: ($\leftarrow$) Suppose BWOC, that there exists an $s$ where $\Pi_s \models q$ and $w \models \nu_s()$ but $w \not\models q$. Consider the rules of the form $\mu_r() \leftarrow \nu_s()$ and $h_r \leftarrow b_r \wedge \mu_r()$ in $\Pi_s$. We know all these rules can also be found in $\Pi_{nil}$. Therefore, as $\Pi_s \models q$, any world that models $\nu_s()$ must also satisfy $q$ - a contradiction.\\

The proof of the lemma follows from claims 1-3.
\end{proof}

\noindent\textit{Proof of Theorem~\ref{theorem:nilson}.}
\begin{proof}
Let $W_s = \{w | w \models \nu_s()\}$. Note that by claim 1 of Lemma 1, that no world (with non zero probability) in $W_s$ can satisfy a helper atom other than $\nu_s()$, so we can partition the set of non-zero probability worlds by helper atoms. Consider the set $W_q \equiv \{ w | w \models q\}$ (the set of worlds we sum over in the left-hand side of the equation). By Lemma 1, there is a precise set of $W_s$'s such that their union equals $W_q$ and that we can characterize those as follows:
\[
W_q = \bigcup_{\Pi_s \models q}W_s
\]
Hence, we have the following equality:
\[
\sum_{w \textit{ s.t. }w\models q}I(w) =\sum_{\Pi_s \models q}\sum_{w \in W_i}I(w)
\]
By the construction, we have the following:
\[
\sum_{w \textit{ s.t. }w\models q}I(w) =\sum_{\Pi_s \models q}\sum_{w \models \nu_s()}I(w)
\]
By step 2 of the construction, we then have:
\[
\sum_{w \textit{ s.t. }w\models q}I(w) =\sum_{\Pi_s \models q}\M(s)
\]
Which gives us the statement of the theorem.
\end{proof}
\section{Algorithm Walkthroughs.}
In \textbf{Algorithm~\ref{alg:learn_pc}}, line 1 constructs the rule-sample association matrix $M$. Line 2 learns the PC topology from $M$. Line 3 fits the PC parameters by EM, treating the missing entries of $M$ as latent variables. 


\noindent In \textbf{Algorithm~\ref{alg:inference}}, line 1 ranks all rules by their marginal probability $\M(\mu_r())$ under the learned PC. Lines 3–7 construct the candidate rule sets for the singleton variants: the top $r_{max}$ rules, each forming a singleton set paired with its marginal. Line 9 instead constructs rule sets via greedy walks (Algorithm 3 in the Appendix) for the GreedyLB variant. Lines 12–14 run standard AnyBURL inference restricted to each selected rule set, collecting the predicted triples. Finally, lines 15–18 assign each predicted triple a probability: for each prediction q, the set $F_q$ of rules active for it is identified, and its probability is computed according to the variant, i.e. the exact value via Proposition~\ref{prop:eq} for SingletonExact, or the lower bound of Proposition~\ref{prop:lb_prop} for SingletonLB and GreedyLB.

\section{Additional Algorithms and Formal Definitions}

\noindent\textbf{Ruleset Generation.}
This section gives algorithms for two different ways of generating rule sets. 
\begin{algorithm}[h]
\caption{Greedy Ruleset Generation for GreedyLB}
\label{alg:greedy}
\textbf{Input}: $\Pi$ (set of learned rules), $P_{\theta}$ (learned distribution), $\delta$ (threshold)\\
\textbf{Output}:\text{ Collection of ordered rule sets $\mathcal{S}$}
\begin{algorithmic}[1] 
\STATE Initialize $R_{\textit{remaining}} = \Pi$ and $\mathcal{S} = \emptyset$.
\WHILE{$R_{\textit{remaining}} \neq \emptyset$}
\STATE $r^* = \arg\max_{r \in R_{\textit{remaining}}} P_\theta(\{r\})$
\STATE Initialize current ruleset $R = \{r^*\}$.
\WHILE{$P_\theta(R) < \delta$}
\STATE $r^* = \arg\max_{r \in R_{\textit{remaining}} \setminus R} P_\theta(R \cup \{r\})$
\STATE $R = R \cup \{r^*\}$
\STATE $\mathcal{S} = \mathcal{S} \cup \{R\}$
\STATE $R_{\textit{remaining}} = R_{\textit{remaining}} \setminus R$
\ENDWHILE
\STATE \textbf{return} $\mathcal{S}$
\ENDWHILE
\end{algorithmic}
\end{algorithm}

\begin{algorithm}[h]
\caption{Singleton Ruleset Generation for SingletonLB and SingletonExact}
\label{alg:singleton}
\textbf{Input}: $\Pi$ (set of learned rules), $P_\theta$ (learned distribution)\\
\textbf{Output}: \text{Collection of ordered singleton rule sets $\mathcal{S}$}
\begin{algorithmic}[1] 
\STATE Initialize $\mathcal{S} = \emptyset$.
\STATE Sort rules in $\Pi$ by marginal probability: $r_1, r_2, \ldots, r_{|\Pi|}$ such that $P_\theta(\{r_1\}) \geq P_\theta(\{r_2\}) \geq \ldots \geq P_\theta(\{r_{|\Pi|}\})$
\FOR{$i = 1$ to $|\Pi|$}
\STATE $R = \{r_i\}$
\STATE $\mathcal{S} = \mathcal{S} \cup \{R\}$
\ENDFOR
\STATE \textbf{return} $\mathcal{S}$
\end{algorithmic}
\end{algorithm}

\begin{theorem}
    The greedy walk algorithm for probabilistic rule selection has time complexity $\mathcal{O}(R \cdot C)$ where $R$ is the number of rules and $C$ is the cost of evaluating marginals over the Probabilistic circuit.
\end{theorem}

\begin{proof}
    The algorithm performs $W$ greedy walks where $W \leq R$. In walk $w$, let $R_w$ denote the number of rules remaining before walk $w$ and $L_w$ denote the number of rules selected in walk $w$. The walks form a partition of all the rules.

    For walk $w$ with $R_w$ initial remaining rules, we analyze the scenarios evaluated in each iteration:
\begin{itemize}
    \item Iteration 1: We start with $R_w$ remaining rules, evaluating $R_w$ scenarios
    \item Iteration 2: After selecting 1 rule, we have $(R_w - 1)$ remaining rules, evaluating $(R_w - 1)$ scenarios
    \item Iteration $i$: After selecting $(i-1)$ rules, we have $R_w - (i-1)$ remaining rules, evaluating $R_w - (i-1)$ scenarios
\end{itemize}

Since walk $w$ performs $L_w$ iterations until probability threshold violation, walk $w$ requires $L_w$ probabilistic circuit queries to be executed.
Aggregating across all walks, the total number of probabilistic circuit queries over all walks is:\\
$$Q_{\text{total}} = \sum_{w=1}^W L_w$$
To get time complexity in worst case, we need to consider a scenario where each walk selects only one rule ($L_w = 1$ $\forall w$). This gives $W = R$ total greedy walks each requiring exactly one probabilistic circuit query. We get total number of queries in worst case as:

\begin{align}
Q_{\text{total}} &= \sum_{w=1}^W 1 \\
&= R
\end{align}

Since each PC query has a fixed computational cost $C$ that depends on the structure of the circuit, the worst case time complexity can be computed as: 

$$T_{\text{total}} = Q_{\text{total}} \times C = \mathcal{O}(R) \times C = \mathcal{O}(R \cdot C)$$

Note that $C$ depends on the structure of the learned Probabilistic Circuit making the overall complexity sensitive to the circuit's computational requirements for each marginal likelihood query.

\end{proof}

\noindent\textbf{Evaluation Metrics.}
This section provides formal definitions for the evaluation metrics used in our experiments. Let $\mathcal{T}_{test}$ denote the set of test triples of particular dataset, and for each test triple $\tau_i \in \mathcal{T}_{test}$, let $rank_i$ denote the rank of the correct entity in the sorted list of predictions (where rank 1 is the highest scoring prediction).

\noindent\textit{Hits@k.} The Hits@k metric measures the proportion of test queries where the correct entity appears within the top-k predictions:
\begin{equation}
\text{Hits@k} = \frac{1}{|\mathcal{T}_{test}|} \sum_{i=1}^{|\mathcal{T}_{test}|} \mathbf{1}[rank_i \leq k]
\end{equation}

\noindent\textit{Mean Rank Reciprocal (MRR).} The MRR metric computes the average reciprocal rank of correct predictions across all test triples. 
\begin{equation}
\text{MRR} = \frac{1}{|\mathcal{T}_{test}|} \sum_{i=1}^{|\mathcal{T}_{test}|} \frac{1}{rank_i}
\end{equation}
Since we are using AnyBURL evaluation engine, which approximates the MRR based on the top-k predictions only, we have used k = 1000 for all our experiments to maintain consistency between the PC framework and baseline. We define MRR as: 
\begin{equation}
\text{MRR} = \text{MRR@k} = \frac{1}{|\mathcal{T}_{test}|} \sum_{i=1}^{|\mathcal{T}_{test}|} \frac{\mathbf{1}[rank_i \leq k]}{rank_i}
\end{equation}
\section{Additional Results}
\begin{figure}[h]
\centering
\includegraphics[width=0.49\columnwidth]{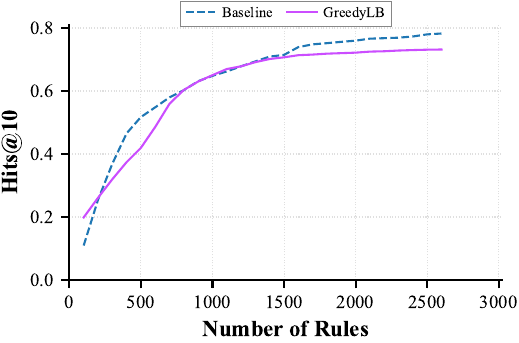} 
\includegraphics[width=0.49\columnwidth]{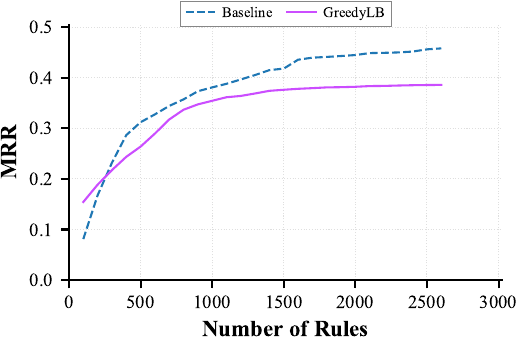} 
\caption{Greedy walks performance on Family dataset. Comparison of GreedyLB and Baseline for Hits@10 (left) and MRR (right).}
\label{fig:Family_greedy}
\end{figure}

\begin{figure}[h]
\centering
\includegraphics[width=0.49\columnwidth]{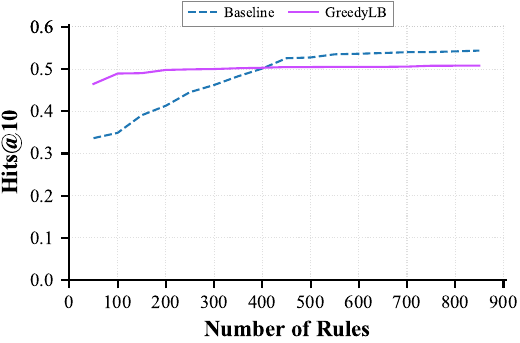} 
\includegraphics[width=0.49\columnwidth]{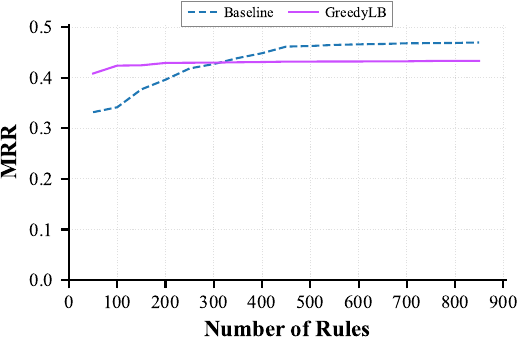} 
\caption{Greedy walks performance on WN18RR dataset. Comparison of GreedyLB and Baseline for Hits@10 (left) and MRR (right).}
\label{fig:WN18RR_greedy}
\end{figure}

\begin{figure}[h]
\centering
\includegraphics[width=0.49\columnwidth]{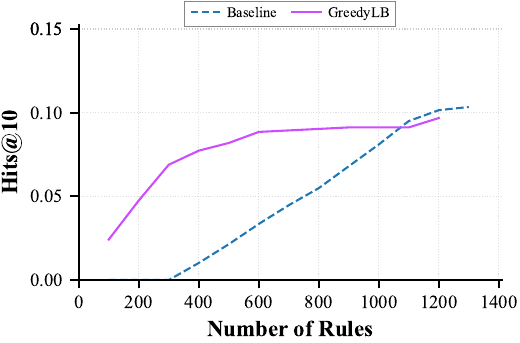} 
\includegraphics[width=0.49\columnwidth]{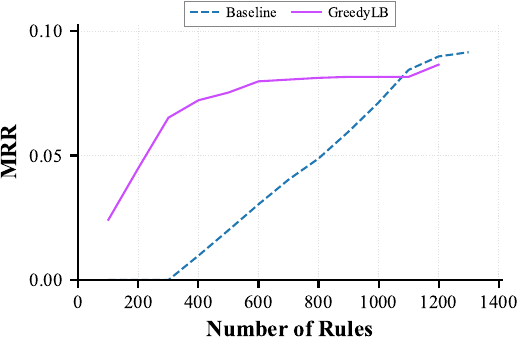} 
\caption{Greedy walks performance on Kinship dataset. Comparison of GreedyLB and Baseline for Hits@10 (left) and MRR (right).}
\label{fig:Kinship_greedy}
\end{figure}

\begin{figure}[h]
\centering
\includegraphics[width=0.49\columnwidth]{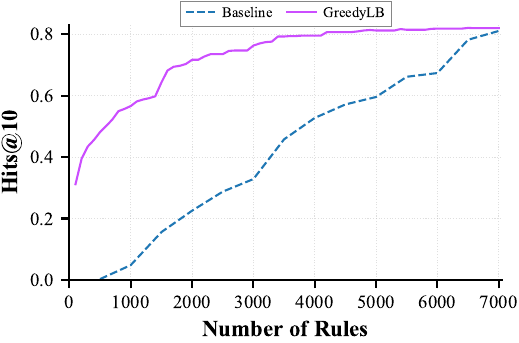} 
\includegraphics[width=0.49\columnwidth]{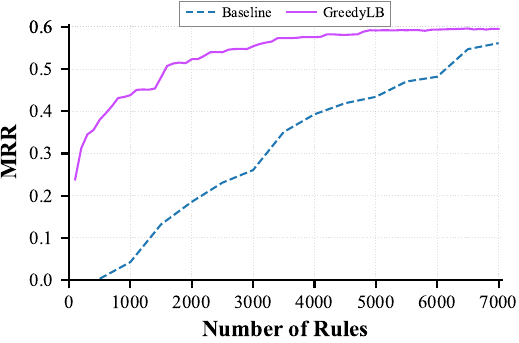} 
\caption{Greedy walks performance on UMLS dataset. Comparison of GreedyLB and Baseline for Hits@10 (left) and MRR (right).}
\label{fig:UML_greedy}
\end{figure}

\begin{figure*}[t!]
\centering
\begin{tabular}{@{}c@{\hspace{0em}}c@{\hspace{0em}}c@{\hspace{0em}}c@{}}

\begin{tabular}{@{}c@{}}
    \textbf{WN18RR} \\[0em]
        \includegraphics[width=0.25\linewidth]{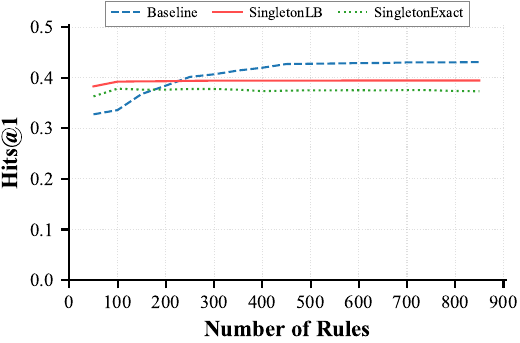}\\[0.8em]

    \textbf{WN18} \\[0em]
        \includegraphics[width=0.25\linewidth]{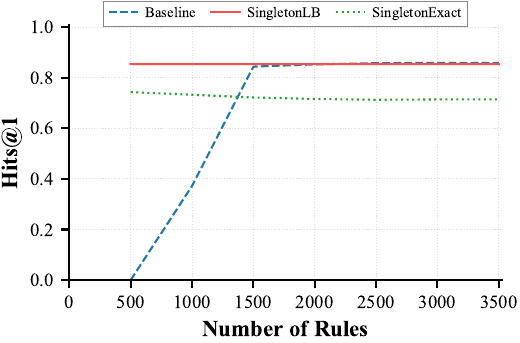}
\end{tabular}
&

\begin{tabular}{@{}c@{}}
    \textbf{Nations} \\[0em]
        \includegraphics[width=0.25\linewidth]{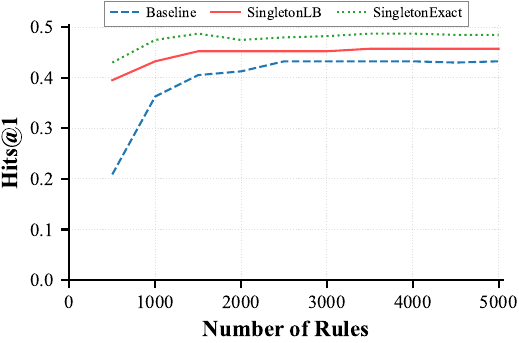}\\[0.8em]

    \textbf{Family} \\[0em]
        \includegraphics[width=0.25\linewidth]{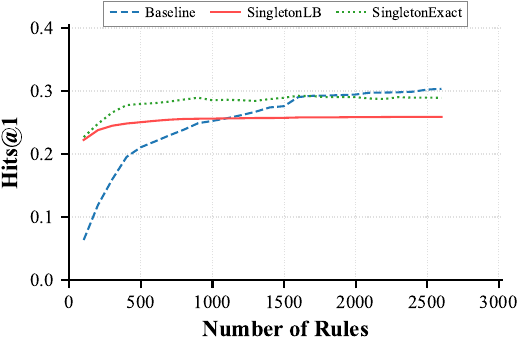}
\end{tabular}
&

\begin{tabular}{@{}c@{}}
    \textbf{FB15K-237} \\[0em]
        \includegraphics[width=0.25\linewidth]{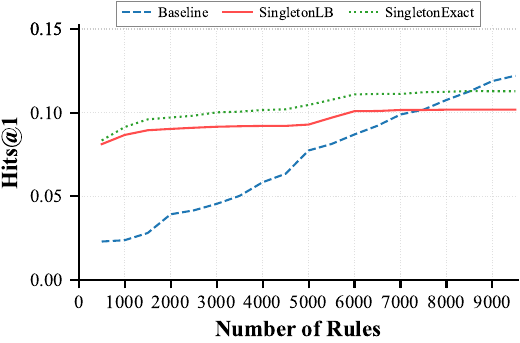}\\[0.8em]

    \textbf{Kinship} \\[0em]
        \includegraphics[width=0.25\linewidth]{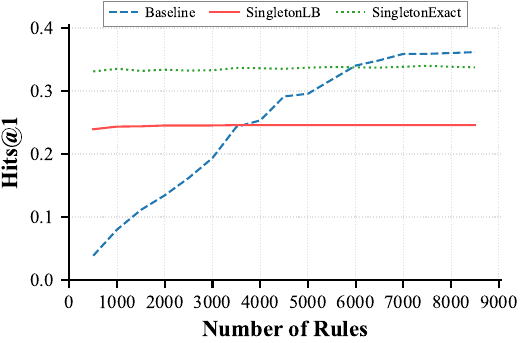}
\end{tabular}

&

\begin{tabular}{@{}c@{}}
    \textbf{CODEX-S} \\[0em]
        \includegraphics[width=0.25\linewidth]{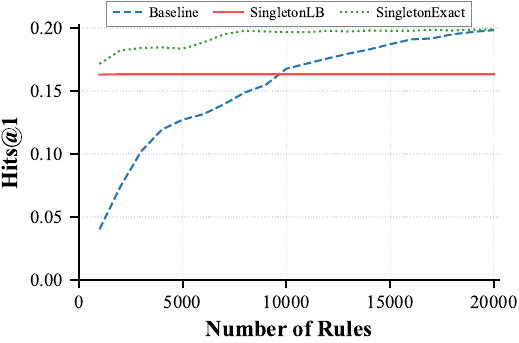}\\[0.8em]

    \textbf{UML} \\[0em]
        \includegraphics[width=0.25\linewidth]{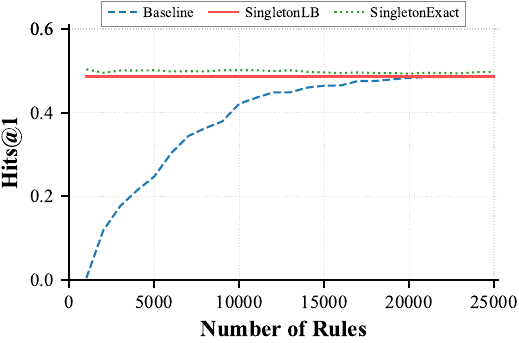}
\end{tabular}

\end{tabular}

\caption{Hits@1 performance of SingletonLB and SingletonExact inferences against Baseline. PC variants attained competitive performance for Hits@1 with reduced rule sets across all evaluated datasets.}
\label{fig:hits_at_1}
\end{figure*}

\begin{figure*}[ht!]
\centering
\begin{tabular}{@{}c@{\hspace{0em}}c@{\hspace{0em}}c@{\hspace{0em}}c@{}}

\begin{tabular}{@{}c@{}}
    \textbf{WN18RR} \\[0em]
        \includegraphics[width=0.25\linewidth]{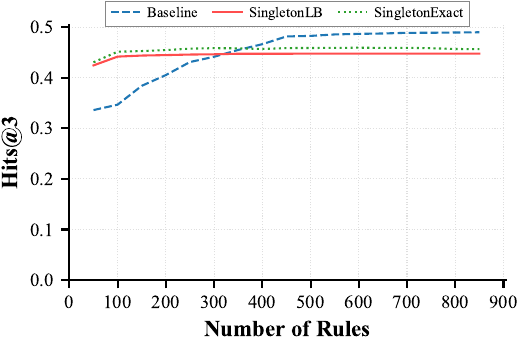}\\[0.8em]

    \textbf{WN18} \\[0em]
        \includegraphics[width=0.25\linewidth]{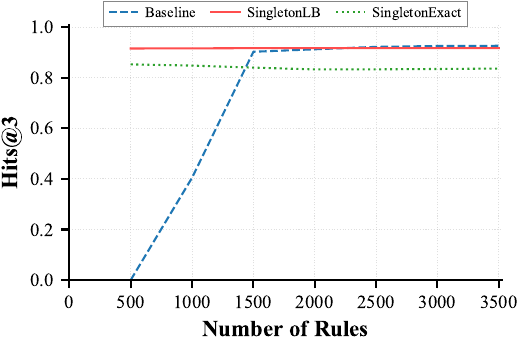}
\end{tabular}
&

\begin{tabular}{@{}c@{}}
    \textbf{Nations} \\[0em]
        \includegraphics[width=0.25\linewidth]{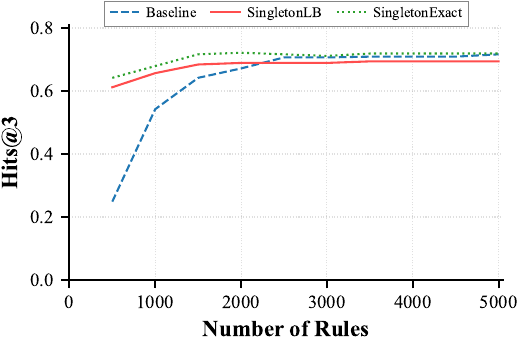}\\[0.8em]

    \textbf{Family} \\[0em]
        \includegraphics[width=0.25\linewidth]{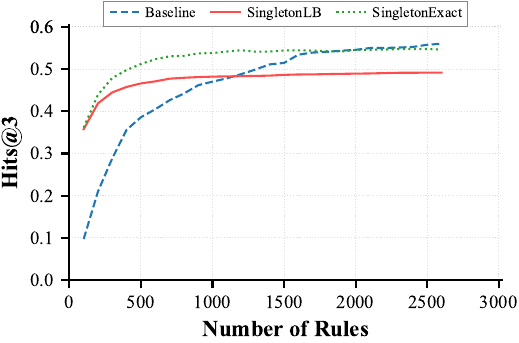}
\end{tabular}
&

\begin{tabular}{@{}c@{}}
    \textbf{FB15K-237} \\[0em]
        \includegraphics[width=0.25\linewidth]{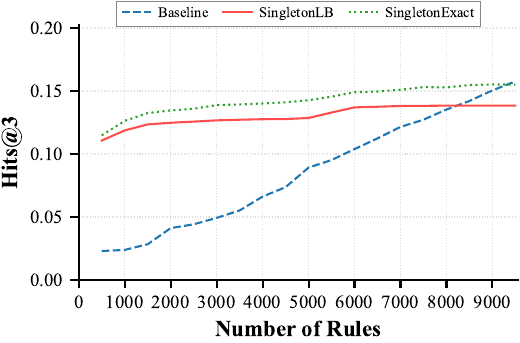}\\[0.8em]

    \textbf{Kinship} \\[0em]
        \includegraphics[width=0.25\linewidth]{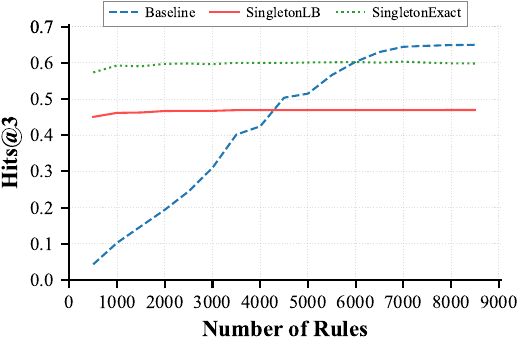}
\end{tabular}

&

\begin{tabular}{@{}c@{}}
    \textbf{CODEX-S} \\[0em]
        \includegraphics[width=0.25\linewidth]{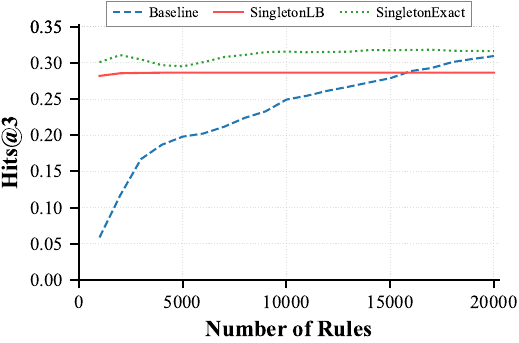}\\[0.8em]

    \textbf{UML} \\[0em]
        \includegraphics[width=0.25\linewidth]{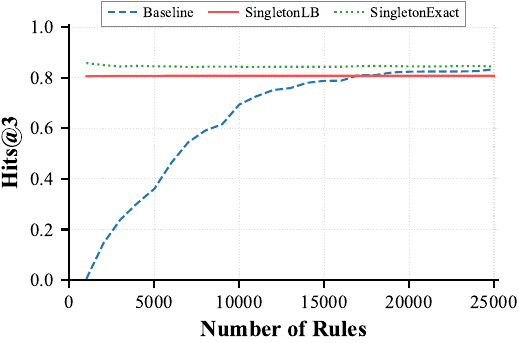}
\end{tabular}
\end{tabular}
\caption{Hits@3 performance of SingletonLB and SingletonExact inferences against Baseline. PC variants attained competitive performance for Hits@3 with reduced rule sets across all evaluated datasets.}
\label{fig:hits_at_3}
\end{figure*}

\begin{figure}[htbp]
\centering
\includegraphics[width=0.48\columnwidth]{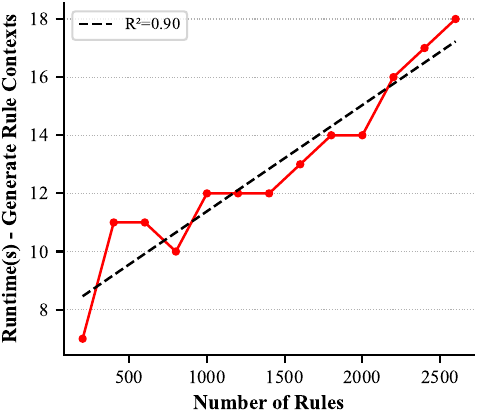} 
\includegraphics[width=0.48\columnwidth]{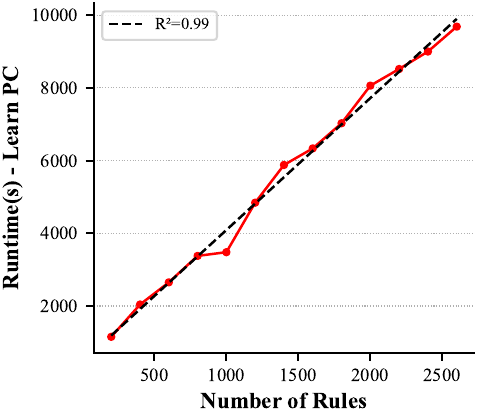} 
\caption{Runtime scalability for generating rule samples (left) and learning PC (right). Runtime for increasing number of non-ground rules - showing linear trend.}
\label{fig:runtime_expand_pc}
\end{figure}

\begin{figure}[htbp]
\centering
\includegraphics[width=0.48\columnwidth]{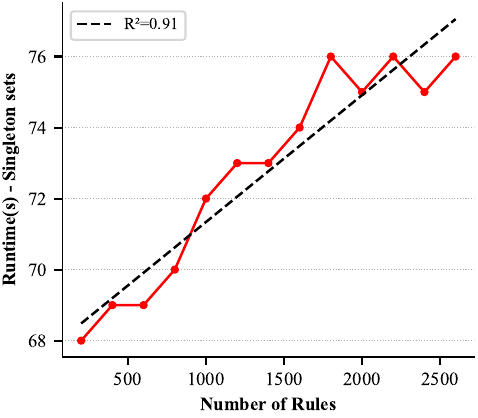} 
\includegraphics[width=0.48\columnwidth]{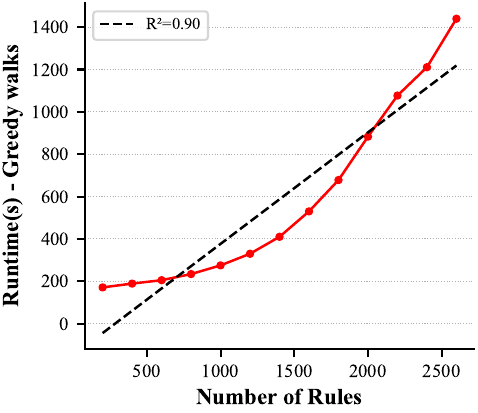} 
\caption{Runtime comparison for singleton(left) vs greedy walks(right) in seconds for increasing number of rules in Family dataset.}
\label{fig:runtime_singleton_greedy}
\end{figure}

\begin{figure}[htbp]
\centering
\includegraphics[width=0.489\columnwidth]{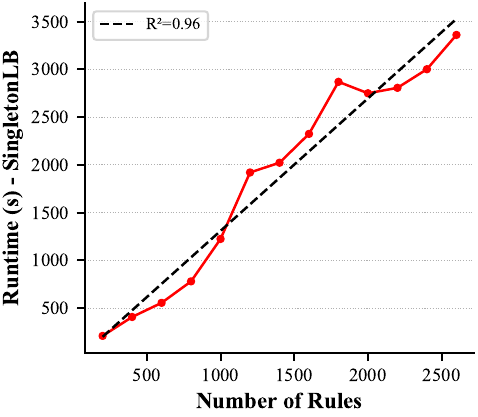} 
\includegraphics[width=0.489\columnwidth]{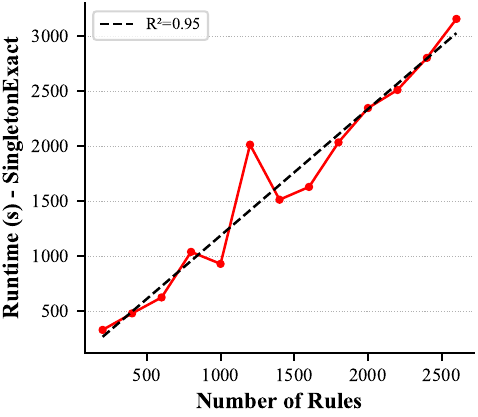} 
\caption{Runtime in seconds of SingletonLB(left) and SingletonExact(right) inferences in seconds for increasing rules in Family dataset.}
\label{fig:runtime_pc1_pc2}
\end{figure}

Figure~\ref{fig:Family_greedy}, Figure~\ref{fig:WN18RR_greedy}, Figure~\ref{fig:Kinship_greedy}, and Figure~\ref{fig:UML_greedy} show GreedyLB (greedy walks) performance compared to baseline across Family, WN18RR, Kinship, and UMLS respectively. This basically extends GreedyLB results shown for Nations datasets in the paper. We see that GreedyLB dominates baseline for smaller number of rules consistently. Note that due to computational complexity of greedy walks, UMLS results use 80\% confidence threshold (compared to 50\% for SingletonLB, SingletonExact) and Kinship uses 90\% confidence threshold (compared to 50\% for SingletonLB, SingletonExact results). WN18RR and Family maintain 0\% confidence threshold as their rule sets remain computationally tractable even with all rules included for GreedyLB. Furthermore, Figure~\ref{fig:hits_at_1} and Figure~\ref{fig:hits_at_3} show SingletonLB and SingletonExact results for Hits@1  and Hits@3 respectively. We see that the results are consistent with the Hits@10 and MRR results shown in the paper for all the datasets. We also see that our framework scales linearly with the number of input rules. For the 100-2600 non-ground rules in the Family dataset (Figure~\ref{fig:runtime_expand_pc}), it took 7-18 seconds to generate rule samples ($R^2=0.90$) and around 1000-9500 seconds to learn the circuit ($R^2=0.99$).
Circuit learning takes most of the computation time here, but enables improved explainability. Figure~\ref{fig:runtime_singleton_greedy} validates that singleton rule sets are much faster: requiring a maximum of 76 seconds, while greedy walks needed around 1500 seconds for 2600 non-ground rules. For larger rule sets (10000 to 15000), greedy walks took more than 24 hours to complete. Figure~\ref{fig:runtime_pc1_pc2} shows that both SingletonLB and SingletonExact scaled linearly ($R^2\ge0.96$), confirming the linear scalability of our inferences.


\section{Code Appendix}
\label{appendix:code}

This section provides an overview of the implementation of our PC-guided inference framework. Note that we provide the summary here and the detailed script descriptions are available in the \texttt{README.md} of the attached PC-Context-Learning directory of the supplementary materials provided. The source code is also accompanied with all the datasets and rule files used.

\noindent\textbf{Framework Architecture.}
Our implementation consists of five major steps executed one after the other:

\begin{enumerate}
    \item Data preprocessing: Generate rule-sample associations using PyClause.
    \item Learning PC: Learn probabilistic circuit structure and parameters via EM.
    \item Generate rule sets: Generate rule subsets using marginal probabilities
    \item Generate predictions: Create SingletonLB, SingletonExact, GreedyLB predictions in AnyBURL format
    \item Evaluation: Compute Hits@k and MRR metrics using AnyBURL evaluation engine for SingletonLB, SingletonExact, GreedyLB, and baseline.
\end{enumerate}

\noindent\textbf{Step 1: Data preprocessing}
\begin{verbatim}
# Generate PyClause groundings
python 1-0-PyClause.py

# Create binary rule-sample matrix  
python 1-1-BinaryMatrixPyclause_script.py 
--dataset Family --confidence 0
\end{verbatim}

\textit{Output}: Binary matrix $\mathbf{M}$ where $\mathbf{M}_{r,s} = 1$ if rule r predicts sample s, missing otherwise.\\

\noindent\textbf{Step 2: PC Learning}
\begin{verbatim}
cd all_scripts/scripts_julia

# Learn PC structure and parameters via EM
julia 2-1-PCbyEM_PyClause.jl

# Compute marginal probabilities using kMAP
julia 2-2-EM-EVI_AnyBURL_KMAP.jl  
\end{verbatim}

\textit{Output}: Learned PC $\mathcal{C}$ and marginal probabilities $P_\theta(\mu_r())$ for each rule $r$.\\

\noindent\textbf{Step 3: Rule Selection}
Two approaches to generate rule subsets are,

\noindent\text{SingletonLB (Singleton Rules):}
\begin{verbatim}
python 2-3-0-NoGreedyWalk.py
\end{verbatim}

\noindent\text{GreedyLB (Greedy Walks):}  
\begin{verbatim}
julia 2-3-RuleSelection_V2.jl
\end{verbatim}

\textit{Output}: Ranked rule subsets each with marginals.\\

\noindent\textbf{Step 4: Prediction}

\noindent\textit{SingletonLB/GreedyLB Predictions:}
\begin{verbatim}
python 3-A-1-GenerateRulesAndPredictions_
script.py
\end{verbatim}

\noindent\textit{SingletonExact Predictions:}
\begin{verbatim}
# Create rule mappings
python 3-B-1-PyClauseMapping_script.py

# Generate predictions  
python 3-B-2-UpperBoundPrediction_script.py

# Calculate negation probabilities
julia 3-B-3-CalculateNegationProbability_
script.jl

# Rewrite with PC scores
python 3-B-4-UpperBoundRewritePrediction_
script.py
\end{verbatim}

\noindent\textit{Baseline Predictions:}
\begin{verbatim}
python 3-C-AnyBurlOnly.py
\end{verbatim}

\noindent\textbf{Step 5: Evaluation}
\begin{verbatim}
# Evaluate baseline
python eval_anyburl_baseline_only.py

# Evaluate SingletonLB/GreedyLB  
python eval_anyburl_pc1_only.py

# Evaluate SingletonExact
python eval_anyburl_pc2_only.py
\end{verbatim}

\textit{Output}: csv file for each SingletonLB, SingletonExact, baseline with Hits@1, Hits@3, Hits@10 and MRR metrics for all $k$.

The exact parameter values to be chosen in the scripts and used for each dataset are specified in the Experimental Setup section of the paper.
Complete implementation details, installation instructions, and usage examples are provided in the project \texttt{README.md} file in the supplementary materials. Note that the implementation pipeline is configured to be executed on Family dataset by default with all the parameter values as specified in the Experimental setup section of the paper.

\end{document}